\documentclass[conference]{IEEEtran}
\IEEEoverridecommandlockouts
\usepackage{cite}
\usepackage{amsmath,amssymb,amsfonts}
\usepackage{algorithmic}
\usepackage[linesnumbered, ruled, lined,boxed,commentsnumbered]{algorithm2e}[1]
\usepackage{graphicx}
\usepackage{textcomp}
\usepackage{xcolor}
\usepackage{multirow}
\def\BibTeX{{\rm B\kern-.05em{\sc i\kern-.025em b}\kern-.08em
    T\kern-.1667em\lower.7ex\hbox{E}\kern-.125emX}}
\begin{document}

\title{DBFed: Debiasing Federated Learning Framework based on Domain-Independent}

\author{\IEEEauthorblockN{Jiale Li}
\IEEEauthorblockA{\textit{School of Software} \\
\textit{Dalian University of Technology}\\
Dalian, China \\
lijiale0928@gmail.com}
\and
\IEEEauthorblockN{Zhixin Li}
\IEEEauthorblockA{\textit{School of Computer Science} \\
\textit{Fudan University}\\
Shanghai, China \\
zxli20@fudan.edu.cn}
\and
\IEEEauthorblockN{Yibo Wang}
\IEEEauthorblockA{\textit{School of Software} \\
\textit{Dalian University of Technology}\\
Dalian, China \\
wangyibo5454@gmail.com}
\and
\IEEEauthorblockN{Yao Li}
\IEEEauthorblockA{\textit{School of Software} \\
\textit{Dalian University of Technology}\\
Dalian, China \\
liyao531224@gmail.com}
\and
\IEEEauthorblockN{Lei Wang$^{\ast}$\thanks{* Lei Wang is corresponding author}}
\IEEEauthorblockA{\textit{School of Software} \\
\textit{Dalian University of Technology}\\
Dalian, China \\
 lei.wang@ieee.org}
 
}

\maketitle

\begin{abstract}

As digital transformation continues, enterprises are generating, managing, and storing vast amounts of data, while artificial intelligence technology is rapidly advancing. However, it  brings challenges in information security and data security. Data security refers to the protection of digital information from unauthorized access, damage, theft, etc. throughout its entire life cycle. With the promulgation and implementation of data security laws and the emphasis on data security and data privacy by organizations and users, Privacy-preserving technology represented by federated learning has a wide range of application scenarios. Federated learning is a distributed machine learning computing framework that allows multiple subjects to train joint models without sharing data to protect data privacy and solve the problem of data islands. However, the data among multiple subjects are independent of each other, and the data differences in quality may cause fairness issues in federated learning modeling, such as data bias among multiple subjects, resulting in biased and discriminatory models. Therefore, we propose DBFed, a \underline{d}e\underline{b}iasing \underline{fed}erated learning framework based on domain-independent, which mitigates model bias by explicitly encoding sensitive attributes during client-side training. This paper conducts experiments on three real datasets and uses five evaluation metrics of accuracy and fairness to quantify the effect of the model. Most metrics of DBFed exceed those of the other three comparative methods, fully demonstrating the debiasing effect of DBFed.
\end{abstract}


\section{Introduction}

Artificial intelligence technology relies heavily on large amounts of data as input, which is used for learning and training to recognize patterns, discover rules, make decisions, and predict outcomes. This data can be structured, such as table data in a database, or unstructured, such as images, text, and speech. Due to the diversity and scale of data, AI technology requires powerful computing capabilities and algorithm support to realize its application value. Additionally, to protect personal privacy and data security, reasonable restrictions and protections need to be placed on the use of sensitive data. Therefore, as AI technology continues to evolve, ensuring the quality and security of data in terms of acquisition, storage, sharing, and utilization becomes an important issue. For enterprises, expanding data sources, establishing a complete data lifecycle management system, and adopting privacy-preserving computing technology can better utilize data resources, improve the efficiency of mining and utilizing data value, and better meet the needs of data security and privacy protection. Therefore, with the development of artificial intelligence technology, how to ensure the quality and security of data in terms of data acquisition, storage, sharing, and utilization through privacy computing technology has become an important frontier research topic. Many fields, such as finance, healthcare, and communication\cite{2021UQCom}, have extremely high requirements for data security. As a result, data between different institutions often becomes a data island, making it difficult to share and use it safely, resulting in ineffective utilization of data value and hindering the development of artificial intelligence technology. In order to protect data security, Privacy-preserving computation technology represented by federated learning has been widely applied in the field of artificial intelligence.

Privacy-preserving computation refers to a computing mode for computing and data processing under the premise of protecting data privacy. It can encrypt, share, calculate, and analyze data without exposing the original data, thereby protecting personal privacy and business secrets. Federated learning is a distributed machine learning technology that allows multiple participants to jointly train a model, but each participant can only access local data, thereby protecting data privacy. Federated learning can avoid collecting and storing user data on a central server, which better protects user privacy. However, the issue of fairness also becomes more important when using federated learning. Fairness issues usually involve inequalities between different participants, including data imbalance, computing resource imbalance, capability differences, etc. Fairness issues in federated learning can be grouped into the following categories:
\subsubsection{Data Bias}The data distribution and characteristics of different participants may differ, leading to the imbalanced performance of the model among different participants. Factors such as gender, race, geographical location, etc. can lead to the neglect or unfair treatment of certain clients' datasets.
\subsubsection{Model Bias}Since different participants have different data distributions and characteristics, the model may be biased towards certain participants, resulting in the unbalanced performance of the model, which will also affect model fairness.
\subsubsection{Imbalance of Computing Resources}Different participants may have different computing resources. Some participants may have more computing power and storage resources to conduct more iterative training locally, while others may only have limited iterative training. This may result in some participants having better model performance than others, leading to fairness issues.
\subsubsection{Capability Differences}Different participants may have different abilities and professional knowledge. Some participants may have more domain expertise and experience to better understand and interpret the model results, while others may lack these abilities, which may lead to issues with the interpretability and fairness of the model.

Addressing bias issues in federated learning can improve the robustness and generalization ability of federated learning models, as well as improve the model's coverage and service quality for different groups. Therefore, in order to address these fairness issues, it is necessary to conduct relevant research and develop appropriate algorithms and tools to promote fair federated learning. This includes methods based on multi-party data joint learning, the use of distributed privacy protection technology in federated learning, and the development of appropriate metrics to evaluate the fairness of the model.

The main contributions of this paper are as follows:
\begin{itemize}
\item This paper proposes a debiasing federated learning framework based on domain-independent, which alleviates model bias by explicitly encoding sensitive attributes during client training, effectively improving the fairness of deep learning classification models in federated learning.
\item This paper conducts experiments on 3 real datasets, and uses 4 fairness indicators to quantify the debiasing effect of the model for multi-classification and multi-sensitive attribute tasks. The effectiveness of DBFed was verified through experiments.
\end{itemize}
 
\section{Related Work}

\subsection{Fedrated Learning}
Federated learning is a distributed machine learning framework proposed by McMahan et al.\cite{2016Communication}, which allows clients to collaborate with each other to train machine learning models without exposing their own data. Nowadays, federated learning has been applied to fields such as healthcare and finance. Yang et al. \cite{2019Federated} and others divided federal learning into horizontal federal learning, vertical federal learning, and federal transfer learning according to the distribution and alignment characteristics of data in the model training process. In the horizontal federated learning framework, datasets from different clients share the same sample space, but the samples are different; In the vertical federated learning framework, datasets from different clients have the same or partially identical sample IDs, but the feature spaces of the datasets are not the same; In the federated transfer learning framework, the sample ID and feature space of datasets on different clients are different, but different clients have similar business scenarios.

In federated learning, the client protects local data privacy by passing model gradients or weights instead of data sharing. However, Zhu et al. \cite{2019Deep} showed that gradient leakage attacks could achieve recovery of client datasets, thereby compromising customer data privacy. After obtaining gradients, this method randomly generates a pair of pseudo features and pseudo labels to perform forward and backward propagation, After deriving the gradient, optimize the pseudo features and pseudo labels by minimizing the distance between the pseudo gradient and the real gradient, so that the pseudo data continuously approaches the original data. Hanchi et al. \cite{2021GRNN} proposed an adversarial generation network for generating random data to minimize the distance between pseudo gradients and real gradients, thereby inferring the raw data of federated learning clients. Melis et al. \cite{2018Inference} have demonstrated that malicious clients can deceive the global model by using multitasking learning, allowing the model to learn more of its desired features and extract more data information from other clients. In order to protect the privacy of clients in the federated learning process, researchers applied homomorphic encryption \cite{2017Privacy} and differential privacy \cite{2017Differentially} to federated learning. Homomorphic encryption applies the encryption algorithm to the process of gradient exchange. The value encrypted by the algorithm is decoded after addition and multiplication, and the result is the same as that of decoding before the operation. Therefore, it can effectively protect data privacy in the gradient exchange process between the client and the server. But homomorphic encryption reduces the computational efficiency of the federated learning framework. Differential privacy effectively protects data privacy by adding noise to the dataset or blurring certain features through generalization methods, making it difficult for attackers to distinguish between samples and recover data. However, due to modifications made to the dataset, differential privacy technology usually needs to balance accuracy and security.

Many researchers have made trade-offs between security, computational efficiency, and accuracy in federated learning, and have improved and optimized the federated learning framework. The FedCG proposed by Wu et al. \cite{2021FedCG}. utilizes conditional adversarial generation networks to achieve a high level of privacy protection while maintaining the computational performance of the model. FedCG has a private extractor and a public classifier on each client, and in the process of weight aggregation, a client generator is used instead of a public extractor. The client knowledge is aggregated through knowledge distillation, protect and the privacy of extractor weights prevents user information leakage while aggregating client knowledge. In the training process of the client, the extractor and classifier are first trained to learn the features of the local dataset, and the output distribution of the extractor is closer to the generation distribution of the generator. Then, the local adversarial generation network, namely the generator, and the discriminator are trained separately from the local data to improve the accuracy of the model. Zhu et al. \cite{2021Data} proposed a method to aggregate client knowledge through knowledge distillation without using additional data. They set up a lightweight generator on the server and used the learned knowledge as induction bias to adjust local training. This method uses fewer communication times and has good generalization ability.
\subsection{Model bias and debiasing}
Although machine learning models have been applied to a wide range of life scenarios such as face recognition and medical image analysis, some models make decisions based on information such as race, gender, and nationality, resulting in algorithmic bias. As Larson et al. \cite{2016How} once pointed out, the COMPAS system has a certain degree of racial biases. The research of Kohavi et al. \cite{1997Scaling} shows that under the same circumstances, the deep neural network usually predicts that male salaries are higher than females. Ashraf et al. \cite{2020Investigating} pointed out that commercial gender classification systems developed by Microsoft, Face++, and IBM have a high recognition error rate for dark-skinned women. Some works have proposed solutions to algorithmic bias. Mehrabi et al.\cite{2021A}divided the solutions to algorithmic bias into three types: preprocessing, in-process, and post-processing. Preprocessing technology eliminates or alleviates algorithm bias by changing the training dataset; in-process technology modifies the machine learning algorithm itself to eliminate or reduce algorithm bias during training; postprocessing technology usually refers that by regarding the trained model as back-box, the label output by the black-box is recalculated according to a new function to eliminate or reduce algorithm bias. In the field of community detection, Mehrabi et al. \cite{2019Debiasing} proposed a community detection method for nodes with low connection attributes, which can alleviate the bias of low-degree nodes. In the field of classifiers, Bilal  et al. \cite{2015Fairness}proposed a fair constraint to prevent classifiers from making predictions related to sensitive attributes in the data, and Kamishima et al. \cite{2012Fairness}  controlled classification accuracy and results by adjusting regularization parameters and the trade-off between fairness. This regularization method is applicable to any probabilistic discriminant model prediction algorithm. In the field of language models, Bordia et al. \cite{2019Identifying} proposed a regularization loss term for language models, which minimizes the projection of the encoder-trained embeddings onto the gender-encoding embedding subspace, effectively alleviating gender bias in language models. In the field of causal inference, Lu et al. \cite{2016A} proposed a framework for discovering and eliminating bias for causal networks, capturing direct and indirect discrimination through the causal effects of protected attributes on decisions passed along different causal paths.

The problems of algorithm bias and model fairness also exist in the field of federated learning. In recent years, some researchers have proposed some debiasing methods under the framework of federated learning. Zhang et al. \cite{2020FairFL}designed a reward mechanism to adjust the training model’s accuracy and fairness, which drives fairness across all demographic groups and addresses the challenges of limited information and limited coordination. The FairBatch framework proposed by Roh et al. \cite{2020FairBatch}, while retaining the standard training algorithm as an internal optimizer, incorporates an external optimizer to equip the internal problem with additional features, implementing adaptively choosing the mini-batch size, so that it will improve the fairness of the model.  This framework can significantly improve the fairness of any pre-trained model through fine-tuning. Papadaki et al. \cite{2019Towards} proposed an algorithm for maximum and minimum fairness in federated learning, where the server requires each client to explicitly share the performance of the model on each race separately.

\begin{figure}[htbp]
\centerline{\includegraphics[width=0.5\textwidth]{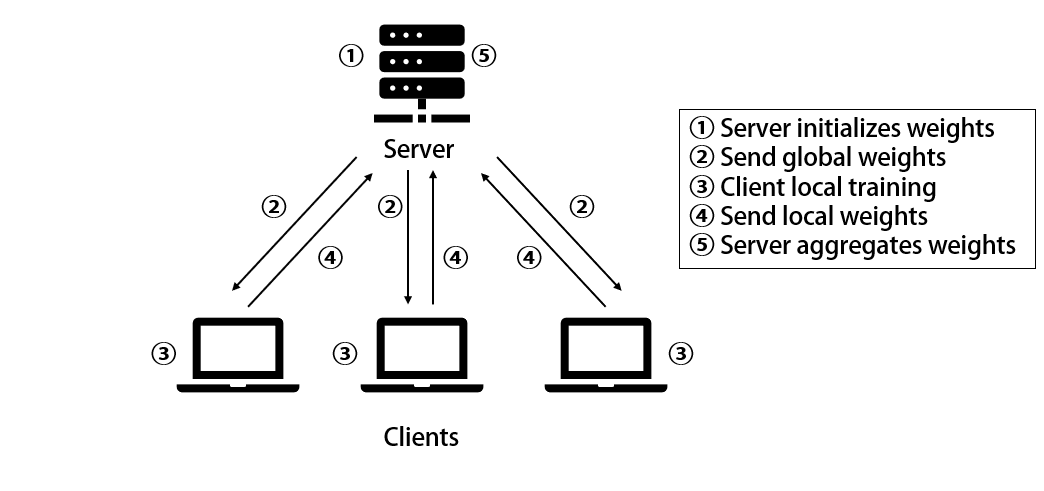}}
\caption{The training process of DBFed.}
\label{fig_3_0_1}
\end{figure}
\section{Method}
This chapter details the federated training process of DBFed. As shown in Figure \ref{fig_3_0_1}, in the learning process of DBFed, it is assumed that there is a global server and clients to jointly train a deep neural network image classification model, the server first initializes the global model and then sends the weights to each client to initialize the local model of the client. After receiving the model weights, the clients use the gradient descent principle to update the model weights on the local dataset to minimize the local loss function. After a certain number of rounds, the clients send the local model weights to the server, and the server performs federated average aggregation on the model weights of the client to obtain global model weights and then distributes the global model weights to the client. The specific process of client training and server aggregation will be introduced in detail below.
\subsection{Client Domain-Independent Training}
Inspired by the research of Wang [35] etc., this paper improves the fairness of the model through domain-independent training, which encoding sensitive attributes explicitly. For the problem of bias in deep learning classification tasks, the predicted features are called target attributes, and the potentially biased populations are called sensitive attributes. The fully connected layer of the deep learning classification model sets $N D$-way discriminant classifiers, where $N$ is the categories of target attribute, and $D$ is the categories of the sensitive attribute. DBFed mitigates model bias by explicitly encoding sensitive attribute information during training and reducing the correlation between sensitive attributes and predicted attributes during prediction. Assuming $f_{}(\cdot)$ is a deep learning classification model, there is $N \times D$ neuron in the last layer of the model, that is, the number output value of the last layer of the model is $N\times D$, recording $f_{z}(x,\theta)$ as the output result of the $z-th$ node of the classification layer, where $x$ is the data sample, and $\theta$ is the model weight. The output of the $N D$-way discriminant classifier can be passed through the activation function:
\begin{equation}
Softmax\left({{f}_{z}}(x;\theta )\right)=\frac{{{e}^{{{f}_{z}}(x;\theta )}}}{\sum\limits_{i=0}^{N}{{{e}^{{{f}_{i}}(x;\theta )}}}}
\end{equation}

The activated data results can be considered as probabilities, then the probability that the predicted result of sample $x$ with sensitive feature $d$ is $y$ is:
\begin{equation}
P (y|d,x)=Softmax (f_{ (y+dN)} (x;\theta))    
\end{equation}

For a data sample, according to the full probability formula, the prediction result of the deep learning classifier can be calculated according to the following formula
\begin{equation}
 \hat{y}=arg{\max \limits_{y}}P(y,x)=arg{\max \limits_{y}}\sum\limits_{d}^{G}{P}(y|d,x)P(d|x)
\end{equation}
where $P(d|x)$ is the probability of sensitive attribute d of the data sample $x$, and $G$ is the set of sensitive attributes, then $|G|=D$. For a data sample with a known sensitive attribute, the predicted value $\hat{y}=arg \max \limits_{y}P(y|d,x)$. However, in order to achieve blind review on the sensitive attribute, that is, to ignore the correlation between predicted attributes and sensitive attributes during the prediction process, as well as achieve Demographic Parity on sensitive attributes, that means, for any $a,b\in G$, $P(a|x)=P(b|x)$, to guarantee the fairness between the prediction different data samples of sensitive attributes and reduce the bias of the algorithm. So in the prediction process, take $P(d|x)=\frac{1}{|G|}$.

For the training data sample x whose real value of the target attribute is $y$ and the sensitive attribute is $d$, its cross entropy can be calculated:
\begin{equation}
L(x,y,d;\theta)=-logP(y|d,x)
\end{equation}
Therefore, using the gradient descent algorithm, the model weight update formula of the client $k$ can be expressed as:
\begin{equation}
{{\theta }^{k}}={{\theta }^{k}}-\eta \nabla \ L\left( b;{{\theta }^{k}} \right) 
\end{equation}
where $\eta$ is the learning rate and $b$ is a batch of training samples. In the local training of the client, the local training data set is divided into multiple batches, and each batch of data is selected for training in each epoch. After multiple iterations of training, the local model weight data is sent to the server.
\begin{algorithm}
    \SetAlgoLined
    \KwIn{
    Communication round $R$, number of clients $K$, number of epoch $E$.
    Training dataset of client $\{ \mathcal{X}_k \}_{k=1}^K$, client local model weights $\{ \theta^k \}_{k=1}^K$.
    Global model weights $\theta_{0}^g$. Learning rate $\eta$ }
    Server initializes $\theta_{0}^g$\;
    \ForEach{round $t$ from 1 to $R$}
    {
        Server sends $\theta_{t}^g$ to clients\;
        //Clients local domain-independent training\;
        \ForEach{client $k$ from 1 to $K$ in parallel}
        {
            Client k receives $\theta_t^g$ \;
            $\theta_1^k=\theta_t^g$\;
            \ForEach{epoch $e$ from 1 to $E$}
            {
                \ForAll{$x,y,d \in \mathcal{X}_k $}
                {
                $\theta_e^k=\theta_e^k-\eta\nabla\ L(x,y,b;\theta_e^k)$\;
                }
                $\theta_{e+1}^k=\theta_{e}^k$\;
            }
            Client sends $k$$\theta_{E}^k$ to Server\;
        }
        //Server aggregates the weights\;
        \textbf{Server executes:}\;
        Server receives$\{\theta_{E}^k \}_{k=1}^K$\;
        $\theta_{t+1}^g=\sum_{k=1}^{K}{\frac{n_k}{n}\theta_t^k}$\;
    }  
    \caption{Debiasing federated learning framework based on domain-independent}
\label{alg1}
\end{algorithm}
\subsection{Server Aggregation}
After the server receives the weight of the client, it performs weight aggregation through the FedAvg algorithm. The aggregation formula of the t+1 round global weight can be expressed as:
\begin{equation}
\theta_{t+1}^g=\sum_{k=1}^{K}{\frac{n_k}{n}\theta_t^k}
\end{equation}
where $n_{k}$ is the number of samples in the local training data set of the client k, and $n=\sum\limits^K\limits_{k=1}n_{k}$ is the sum of the number of samples in the training data set of all clients.
\begin{table*}[htbp]
\centering
\caption{Data Distribution and Setting}
\begin{tabular}{ccccccc}
\hline
\multirow{2}{*}{Dataset} & \multicolumn{2}{c}{Dataset Size} & \multicolumn{2}{c}{Target Attribute}                                      & \multicolumn{2}{c}{Sensitive Attribute}                                 \\
                         & Training Set    & Testing Set    & Label   & \begin{tabular}[c]{@{}c@{}}Number of \\ Categories\end{tabular} & Label & \begin{tabular}[c]{@{}c@{}}Number of \\ Categories\end{tabular} \\ \hline
CelebA                   & 162,770         &  19,962        & Smiling & 2                                                               & Male  & 2                                                               \\
FairFace                 & 86,744          & 10,954         & Age     & 9                                                               & Race  & 7                                                               \\
UTKFace                  & 18,964          & 4,741          & Age     & 9                                                               & Race  & 4                                                               \\ \hline
\label{table1}
\end{tabular}
\end{table*}
\subsection{Joint Training}
In the training process of the federated learning debiasing framework, as shown in Algorithm \ref{alg1}, the global model weights $\theta_{0}^{g}$ are first initialized by the server, and then many rounds of communication are performed. In each round, the server first sends the global model weights of this round to each client, and then each client receives the global model weights and performs local training in parallel. The client executes many iterations for each local training. In the iterations, the batch data sets are extracted in batches, and the local model weights are adjusted through gradient descent to optimize the local model loss function. After the client is trained locally, the client sends the local weights to the server. The server receives the weights of all clients in this round and starts to aggregate the weights of this round to obtain the global model weights of a new round, finally ending the communication training of this round.

\section{Experiment}
\subsection{Dataset}
\subsubsection{CelebA}
CelebA dataset \cite{2016Deep} is a face attribute dataset provided by Liu et al. It contains 202,599 face pictures of 10,177 celebrity identities. The training data set contains 162,770 pictures, the test dataset contains 19,962 pictures, and each picture is marked with 40 features such as gender, hair color and lips. This dataset is widely used in computer vision deep learning tasks.TABLE \ref{table1} shows the data distribution and settings of the dataset in this paper. In the experiment, this paper chooses the "Smiling" label as the target attribute and uses the "Male" label as the sensitive attribute to study the bias of the model in the gender population when predicting smiles.
\subsubsection{FairFace}
The FairFace dataset \cite{2021FairFace} is a face image dataset proposed by Karkkainen et al. It consists of 108,501 pictures, of which the training data set contains 86,744 pictures, and the verification data set contains 10,954 pictures. It includes seven ethnic groups: White, Black, Indian, East Asian, Southeast Asian, Middle Eastern, and Latin. Each facial image is labeled with race, gender, and age, with age attributes classified into nine categories based on age group. In the experiment, this paper chooses the "Age" label as the target attribute and uses the "Race" label as the sensitive attribute to study the bias of the deep learning model on the racial group when predicting age.
\subsubsection{UTKFace}
The UTKFace dataset \cite{2017Age} is a facial dataset with a long age span proposed by Zhang et al., which contains over 20,000 images. In this paper, the dataset is divided into a training dataset with a size of 18,964 and a testing dataset with a size of 4,741 in a ratio of 80\% and 20\%. Each image in the dataset is labeled with gender, age, and race. There are five types of race labels, including white, black, Asian, Indian, and Others. This article uses images from the first four races for experiments\footnote{The number of the last race "Others" is too small and not properly labeled, which has a significant random impact on the experiment.}, using the "Race" label as a sensitive attribute. The age tags in the image are divided into nine categories based on the age groups of "less than 2", "3-9", "10-19", "20-29", "30-39", "40-49", "50-59", "60-69", and "more than 70", with the "Age" label as the target attribute.

\subsection{Evaluating Metrics}
In order to quantify the actual debiasing effectiveness of deep learning classification models, this paper selects one metric to measure classification accuracy and four metrics to measure model fairness as evaluating metrics.
\subsubsection{Accuracy}
Accuracy refers to the probability that the model correctly classifies predictive attributes, which can be calculated by the following formula:
\begin{equation}
ACC=P\left(\hat{Y}=c \middle|\ Y=c\right)    
\end{equation}
where $c\in C$ and $C$ are the set of the target attribute.

\subsubsection{Skewed Error Ratio(SER)}
The skewed error ratio is a metric that evaluates the maximum difference between different sensitive attributes. It mainly represents the difference between the sensitive attribute with the highest accuracy and the sensitive attribute with the lowest accuracy. The larger the value, the greater the difference in the algorithm's discrimination accuracy for different races. The formula can be expressed as:
\begin{equation}
SER=\frac{\min \limits_{g\in G}{Error_g}}{\max \limits_{g\in G}{Error_g}}
\end{equation}
where $g$ is the sensitive attribute, $G$ is the set of sensitive attribute, and $Error_g$ representing the classification error rate of images with sensitive attribute $g$.

\subsubsection{Equal of Opportunity(EO)}
Equal of Opportunity is a metric that evaluates the equal discrimination between different races, mainly indicating the equality of correctly classified races. The larger the value, the greater the difference in the probability of correctly classified races. Achieving equal opportunities requires the model to have an equal true positive or false negative rate, and the conditions for achieving equal opportunities can be expressed as: for all $a,b \in G$, 
$
P\left( \hat{Y}=1 \middle|\ S=a,Y=1 \right)=P\left( \hat{Y}=1\middle|\ S=b,Y=1 \right)
$. 

Since this paper focuses on the situation of multiple categories of target attribute and sensitive attribute, the calculation formula for Equality of Opportunity is defined as the mean accuracy variance of each sensitive attribute image in different target attribute images, as follows:
\begin{equation}
EO=\frac{1}{\left|C\right|}\sum_{c\in C}{\mathop{var}\limits_{g\in G}\left(P\left(\hat{Y}=c\middle|\ S=g,Y=c\right)\right)}    
\end{equation}
where $var(\cdot)$ is variance calculation function.

\subsubsection{Bias Amplification(BA)}
Bias amplification is metric that evaluates the degree of inclination of algorithm decisions towards specific types of target attributes, mainly indicating the unfairness of the algorithm among the types of target attributes. The larger the value, the more inclined the algorithm is towards certain specific target attributes. Bias Amplification can be calculated using the following formula:
\begin{equation}
BA=\frac{1}{\left|C\right|}\sum_{c\in C}\frac{\max \limits_{g\in G}{g_c}}{\sum  \limits_{g\in G}\ g_c}-\frac{1}{\left|G\right|}
\end{equation}

\subsubsection{Demographic Parity(DP)}
Demographic Parity is a metric that evaluates the similarity of algorithm decisions to different races, mainly indicating the degree of similarity of algorithm decisions to different races. The larger the value, the greater the difference in algorithm decisions for different populations. The condition that the model meets Demographic Parity is that all $a,b \in G$, meet $P\left(\hat{Y}=1\middle|\ S=a\right)=P\left(\hat{Y}=1\middle|\ S=b\right)$. The calculation formula for Demography Parity is as follows:
\begin{equation}
DP=\frac{1}{ C }\sum\limits_{c\in C}{\mathop{var}\limits_{g\in G}}\left( P\left( \hat{Y}=c \middle|S=g \right) \right)
\end{equation}
\subsection{Comparative Experiment}
\subsubsection{Environment Settings}
The experimental operating system is Linux 3.10.0, and the development environment is Anaconda3, Python3.10.9, and Pycharm. The deep learning model is mainly written based on the deep learning framework Pyorch 1.13.1 and trained on NVIDIA A100.
\subsubsection{Comparison Methods}
This paper chooses the Federated Average (FedAvg) and local training algorithm as the baseline and chooses Fair Federated Learning Model\cite{2021FairFed} (FairFed) proposed by Ezzeldin et al. as state-of-the-art (SOTA). In local training algorithms, each client only trains on the local dataset and does not aggregate weights through communication.
\subsubsection{Parameter settings}
This paper uses Resnet34 \cite{2016Deep} as the basic model for deep learning for experiments. Resnet34 is a deep residual network with 34 convolutional layers, which is easy to optimize and widely used in computer vision tasks. In the experiment, the adaptive moment estimation (Adam)\cite{2014Adam} optimizer was used to implement the gradient descent principle, which can achieve high computational efficiency in small memory. The learning rate of the optimizer is 0.0001, and the weight decay is 0.0003. During the model training process. Five clients are set up and randomly divided into equal datasets as their local training dataset. The batch size is 128, and the client performs communication aggregation after every three local training iterations. This paper mainly uses three image datasets, CelebA, FairFace, and UTKFace for experiments. Each image has three channels, R, G, and B, with values ranging from 0 to 255. All images are uniformly adjusted to pixel size.
\subsubsection{Results and Analysis}
TABLE \ref{table2} shows the experiment results, the best-performing data for each metric in each dataset are highlighted in bold, while the second-best data is underlined. The experimental effect of DBFed on the CelebA data set is the best. It outperforms other methods in terms of Equality of Opportunity and Demographic Parity and ranks second in terms of Accuracy and Skewed Error Ratio. This shows that DBFed has a good effect on model fairness while having a high prediction accuracy, and effectively reduces the gender bias in smile classification prediction. In the experiments on the FairFace dataset, DBFed achieved the highest Accuracy, but the other four fairness metrics did not significantly surpass other methods. This may be due to the lack of significant unbiased effects caused by the large variety of sensitive attributes in the dataset. In the experiment of UTKFace, this framework outperformed other methods in terms of Skewed Error Ratio, Bias Amplification, and population equality while achieving high accuracy, indicating excellent results in model accuracy and fairness.
Overall, DBFed performs well in both model accuracy and fairness, effectively reducing population discrimination when performing classification tasks. 
\begin{table}[htpb]
\centering
\caption{Experimental result data}
\begin{tabular}{crcccc}
\hline
                          & \multicolumn{1}{l}{Metric} & FedAvg  & Local   & DBFed     & FairFed \\ \hline
\multirow{5}{*}{\rotatebox{90}{CelebA}}   & ACC$\uparrow$                   & 0.9054  & 0.8968  & \underline{0.9058} & \textbf{0.9071}  \\
                          & SER$\downarrow$                   & 1.2673  & \textbf{1.2265}  & \underline{1.2565}   & 1.2607  \\
                          & EO$\downarrow$                      & 0.00044 & 0.00048 & \textbf{0.00022}  & \underline{0.00033} \\
                          & BA$\downarrow$                       & 0.1140  & \textbf{0.1133}  & 0.1156   & \underline{0.1138 } \\
                          & DP$\downarrow$                       & 0.01863 & 0.01855 & \textbf{0.01748}  & \underline{0.01832} \\ \hline
\multirow{5}{*}{\rotatebox{90}{FairFace}} &ACC$\uparrow$                     & 0.5447  & 0.5360  & \textbf{0.5551}   & 0.5485  \\
                          & SER$\downarrow$                       & 1.1315  & \underline{1.1424}  & 1.1214   & \textbf{1.1130}  \\
                          & EO$\downarrow$                       & \underline{0.00461} & 0.00483 & 0.00515  & \textbf{0.00460} \\
                          & BA$\downarrow$                      & \textbf{0.0527}  & 0.0556  & 0.0591   & \underline{0.0546}  \\
                          & DP$\downarrow$                      & \textbf{0.00175} & 0.00190 & 0.00201  & \underline{0.00186} \\ \hline
\multirow{5}{*}{\rotatebox{90}{UTKFace}}  & ACC$\uparrow$                   & 0.5670  & 0.5675  & \underline{0.5727}   & \textbf{0.5783}  \\
                          & SER$\downarrow$                     & \underline{1.5561}  & 1.6297  & \textbf{1.5242}   & 1.668   \\
                          & EO$\downarrow$                      & \textbf{0.00640} & \underline{0.00830} & 0.00997  & 0.0089  \\
                          & BA$\downarrow$                      & 0.3085  & 0.3109  &\textbf{ 0.3001}   & \underline{0.3063}  \\
                          & DP$\downarrow$                       & \underline{0.0418}  & 0.0420  & \textbf{0.0391}   & 0.0422  \\ \hline
\end{tabular}
\label{table2}
\end{table}

\section{Conclusion}
This paper proposes a debiasing federated learning framework
based on domain-independent that can predict classification without using sensitive attribute labels, mitigate model bias model biases during federated learning, and improve model fairness. This paper verifies the depolarization effect of DBFed through experiments on three datasets and five evaluation metrics. In addition, due to the need for sensitive attribute labels during the training process, there are certain requirements for dataset annotation. The research in this paper is a new attempt at model fairness under the federated learning computing architecture, which can be applied to many scenarios that require high data security, such as model fine-tuning and model unbinding in human recognition or fund trading intelligent dialogue models in the financial field. The next research direction will focus on exploring how to remove model biases in federated learning processes without using sensitive attribute labels and strive to perform better debiasing effects on more categories of the sensitive attribute.
based on domain-independent that can predict classification without using sensitive attribute labels, mitigate model bias model biases during federated learning, and improve model fairness. This paper verifies the depolarization effect of DBFed through experiments on three datasets and five evaluation metrics. In addition, due to the need for sensitive attribute labels during the training process, there are certain requirements for dataset annotation. The research in this paper is a new attempt at model fairness under the federated learning computing architecture, which can be applied to many scenarios that require high data security, such as model fine-tuning and model unbinding in human recognition or fund trading intelligent dialogue models in the financial field. The next research direction will focus on exploring how to remove model biases in federated learning processes without using sensitive attribute labels and strive to perform better debiasing effects on more categories of the sensitive attribute.

\section{Acknowledgements}
Joint Research and Development Project of Yangtze River Delta Region Technology and Innovation Community(2022CSJGG0800).

\bibliographystyle{IEEEtran}
\bibliography{Section/references}

\end{document}